\documentclass[conference]{IEEEtran}
\IEEEoverridecommandlockouts
\usepackage{graphicx}
\usepackage{latexsym}
\usepackage{bm}
\usepackage{amssymb}
\usepackage{amsmath}

\newtheorem{definition}{\textbf{Definition}}

\usepackage[table,xcdraw]{xcolor}
\usepackage{multirow}
\usepackage{stfloats}
\usepackage[ruled, vlined, linesnumbered]{algorithm2e}
\usepackage{booktabs}
\def\BibTeX{{\rm B\kern-.05em{\sc i\kern-.025em b}\kern-.08em
    T\kern-.1667em\lower.7ex\hbox{E}\kern-.125emX}}

\makeatletter
\def\ps@IEEEtitlepagestyle{
	\def\@oddfoot{\mycopyrightnotice}
	\def\@evenfoot{}
}
\def\mycopyrightnotice{
	{\footnotesize Under review\hfill} 
	\gdef\mycopyrightnotice{}
}

\begin{document}

\title{Robustness Testing for Multi-Agent Reinforcement Learning: State Perturbations on Critical Agents\\
}

\author{\IEEEauthorblockN{1\textsuperscript{st} Ziyuan Zhou}
\IEEEauthorblockA{\textit{Department of Computer Science} \\
\textit{Tongji University}\\
    China, Shanghai }
\and
\IEEEauthorblockN{2\textsuperscript{nd} Guanjun Liu}
\IEEEauthorblockA{\textit{Department of Computer Science} \\
\textit{Tongji University}\\
    China, Shanghai }
}

\maketitle

\begin{abstract}
Multi-Agent Reinforcement Learning (MARL) has been widely applied in many fields such as smart traffic and unmanned aerial vehicles. However, most MARL algorithms are vulnerable to adversarial perturbations on agent states. Robustness testing for a trained model is an essential step for confirming the trustworthiness of the model against unexpected perturbations. This work proposes a novel Robustness Testing framework for MARL that attacks states of Critical Agents (RTCA). The RTCA has two innovations: 1) a Differential Evolution (DE) based method to select critical agents as victims and to advise the worst-case joint actions on them; and 2) a team cooperation policy evaluation method employed as the objective function for the optimization of DE. Then, adversarial state perturbations of the critical agents are generated based on the worst-case joint actions. This is the first robustness testing framework with varying victim agents. RTCA demonstrates outstanding performance in terms of the number of victim agents and destroying cooperation policies.
\end{abstract}


\section{Introduction}
Multi-Agent Reinforcement Learning (MARL) is wildly utilized in Multi-Agent Systems (MAS) such as smart transportation \cite{9103316} and unmanned aerial vehicles \cite{9993797, 9209079} as a result of its superior performance in team decision-making problems. 
As the number of agents increases and the joint-action space of MAS grows exponentially, MARL faces the issues of high combinatorial complexity and poor scalability. Centralized Training and Decentralized Execution (CTDE) is currently the more popular framework to address these problems, in which, all  agents share the global information in the training process, while in the execution phase, each agent makes independent decisions based on its own perceptions and policy. Value decomposition networks (VDN) \cite{vdn} and monotonic value function factorization (QMIX) \cite{qmix} are classical CTDE-based MARL. They introduce a network in the training process to guarantee the Individual-Global-Max (IGM) \cite{pmlr-v97-son19a}. 

However, it is shown that the agents trained by these classical MARL are sensitive to state perturbation \cite{Guo_2022_CVPR, he2023robust, 9283830, zhou2022romfac}. 
In reality, the states of agents are often perturbed due to the presence of sensor noise and malicious attacks. Furthermore, state perturbations to some of the agents can not only mislead the decision of victims but also have an impact on the cooperative policy of the team. 
Robustness testing for a model trained by MARL is an essential step for confirming the trustworthiness of MAS against unexpected perturbations.
The perturbation types are so diverse that it is impossible to cover all possible cases during testing. Consequently, it is vital to generate perturbation states via adversarial attacks that have the most significant impact on team collaboration. Intuitively, the more stealthy and disruptive the attack is, the more potential it is for robustness testing.


There has been a lot of robustness testing technique on Single-Agent Reinforcement Learning (SARL), such as using the adversarial attack based on the gradient of the neural network \cite{huang2017adversarial, NEURIPS2020_f0eb6568} or constructing an adversary as an RL agent \cite{zhang2021robust} to generate the adversarial observation. 
However, there have been a few related research to test the robustness of MARL against state perturbation of agents. Zhou et al. \cite{zhou2022romfac} demonstrate that the adversary for MARL can be formulated as the Stochastic Game (SG) and there exists the joint optimal adversarial state. But the influence of individual policy on teams is not considered during the attack in \cite{Guo_2022_CVPR, zhou2022romfac}. They only generate adversarial observation misleading the victim to take actions that are not within expectations, which may not lead to the failure of team tasks.
The methods in \cite{han2022solution, li2023attacking, 9283830} consider the effect of individual actions on teams by constructing the adversary as SARL or MARL agent. However, the adversary is trained at the assumption that the victim is determined. When the victim changes, it needs to be retrained. 

Compared with single-agent situations, multi-agent situations face the following challenges: 
\begin{itemize}
    \item[1)]  The victims are uncertain which makes the robustness testing process can not formulate as SG. The testing process becomes more difficult due to this uncertainty.%
    \item[2)]  An agent doing a sub-optimal action does not necessarily lead to team failure.
    \item[3)]  The centralized training process is usually unknown and not be employed during testing. Besides, centralized training is based on the assumption that all agents make the optimal decision. As a result, it may not evaluate accurately the team reward when the agent takes the sub-optimal action. How to estimate the team reward for the adversary is important.
\end{itemize}
Facing these challenges,  this work aims to propose a stealthy and effective attack method on states of Critical Agents for Robustness Testing (RTCA) which has the following novel contributions:
\begin{itemize}
    \item [1)] We propose a novel robust testing framework that has two steps. The first step based on Differential Evolution (DE) aims to select critical agents as victims and determine the worst joint actions they should take to decrease the accumulated reward of the team. The second step is a targeted attack to compute adversarial observation according to the outcomes of the first step.
    \item [2)] We introduce a Sarsa-based approach for learning the joint action-value function MARL to evaluate the team cooperation policy. The function has a good representation of the relationship between individual actions and team accumulated rewards and is used as the objective function for the optimation of DE.
    \item [3)] The results demonstrate that RTCA achieves superior performance when attacking a smaller number of agents. RTCA is more suitable for robustness testing of models trained by MRAL due to its stealthiness and effectiveness.
\end{itemize}

\section{Preliminary}
In this section, we introduce Decentralized Partially Observable Markov Processes (Dec-POMDPs) and State-Adversiral Stochastic Games (SASG).  
Important concepts are described as follows:
\begin{itemize}
    \item Adversarial perturbation aims to mislead the actions of agents to minimize the expected cumulative discount reward of the team.
    \item Clean observation is generated from the environment state. 
    \item Adversarial observation is an observation after adding an adversarial perturbation to a clean observation.
    \item The adversary aims to generate the adversarial perturbation.
    \item Victim is the agent that observation is perturbed by the adversary, i.e., attacked by the adversary.
    \item Critical agents are those whose actions have the greatest impact on the team at a time step. 
\end{itemize}
\subsection{Decentralized Partially Observable
Markov Processes} 
A Dec-POMDP~\cite{oliehoek2016concise} is defined as a tuple 
$$\left<\mathcal{S}, \{\mathcal{A}^i\}_{i\in\mathcal N},\{\mathcal{O}^i\}_{i\in\mathcal N}, \mathcal{N}, Z, p, r, \gamma\right>$$
where $\mathcal{N}$ is the set of agents, the number of agents is $N \mathop=\limits^\Delta |\mathcal{N}|$, each agent $i\in\mathcal{N}\mathop  = \limits^\Delta \{1,\cdots, N\}$, $\mathcal S$ is the state set of the environment, the state $s\in\mathcal{S}$, $a^i \in \mathcal{A}^i$ is the action of agent $i$, $\mathcal A^i$ is the action space of the agent $i$, $o^i \in \mathcal{O}^i$ drawn according to observation function $Z\left(s, \bm a\right): \mathcal{S}\times\mathcal{A}^1\times\cdots\times\mathcal{A}^N\rightarrow\mathcal{O}^1\times\cdots\times\mathcal{O}^N$ is the observation of agent $i$, $\mathcal{O}^i$ is the observation space of agent $i$, $r$ is the immediate reward, for each agent $r^1=\cdots=r^N=r$,
$R\left(s, \bm a, s'\right): \mathcal S \times \mathcal A^1 \times \cdots \times \mathcal A^N \times \mathcal S \rightarrow \mathbb R $ is the reward function of all agents, $p:\mathcal S \times \mathcal A^1 \times \cdots \times \mathcal A^N  \rightarrow \Delta\left(\mathcal S\right)$ is the transition probability based on the joint action $\bm a$ and $\gamma \in \left[0,1\right]$ is the discount factor over time. The environment is the partial observation, thus, the observation of the agent is equal to the agent state in this paper.

It is NEXP-complete to solve Dec-POMDPs due to the combinatorial problem \cite{bernstein2002complexity}. CTED is a potential learning paradigm to solve Dec-POMDPs. There has a centralized controller to evaluate the team cooperation policy based on the environmental state in the training process. And in the execution process, each agent makes action according to its observation. One of the most classical solutions is via Q-function factorization, including VDN \cite{vdn} and QMIX \cite{qmix}. The factorization needs to satisfy the IGM condition \cite{pmlr-v97-son19a}, i.e.,
\begin{equation}
\arg \mathop {\max }\limits_{\bm a}  {Q_{jt}}\left( {\bm \tau ,\bm a} \right) = 
\left( \begin{array}{c}
\arg \mathop {\max }\limits_{a^1}  {Q_{1}}\left( {\tau^1 ,a^1} \right)\\
 \cdots\\
\arg \mathop {\max }\limits_{a^N}  {Q_{N}}\left( {\tau^N ,a^N} \right)
\end{array} \right),
\end{equation}
where $\bm \tau \in \mathcal{T}^1\times\cdots\times\mathcal{T}^N$ is the joint action-observation histories, $\mathcal{T}^i$ is the set of action-observation histories of agent $i$, $Q_{jt}: \mathcal{T}^1\times\cdots\times\mathcal{T}^N\times\mathcal A^1\times\cdots\times \mathcal A^N\rightarrow\mathbb R$ is the joint action-value function to evaluate the team cooperation policy. 

VDN and QMIX are methods to provide additivity and monotonicity two sufficient conditions for IGM, respectively:
\begin{equation}
    Q_{jt}\left(\bm \tau, \bm a\right) = \sum^N_{i=1}Q^i\left(\tau^i, a^i\right),
\end{equation}
\begin{equation}
    \frac{\partial Q_{jt}\left(\bm \tau, \bm a\right)}{\partial Q^i\left(\tau^i, a^i\right)}\ge 0, 0\le i < N .
\end{equation}
\subsection{State-Adversiral Stochastic Game}
An SASG \cite{zhou2022romfac} is defined as a tuple $$\left<\mathcal S, \{\mathcal A^i\}_{i\in \mathcal{N}},\{\mathcal B^i\}_{i\in \mathcal{M}}, \{r^i\}_{i\in \mathcal{N}}, \mathcal{M}, \mathcal{N}, p,\gamma \right>$$ where $\mathcal B^i$ is the set of adversarial states of agent $i$, $\mathcal{M} \subseteq \mathcal{N}$ is the set of victim agents and the number of victims is $M\mathop=\limits^\Delta|\mathcal{M}|$.
Assume that the adversarial perturbation $v^i(s)$ of agent $i$ is deterministic  $v^i:\mathcal S \rightarrow \mathcal B^i$.  
The value and action-value functions of SASG are
\begin{equation} 
\widehat V_{\bm \pi \circ \bm v}^i\left(s\right)=\mathbb E_{\bm \pi \circ \bm v,p}\left(\sum_{k=0}^\infty \gamma^kr^i_{t+k+1}|s_t=s\right),
\end{equation}
\begin{equation}
\widehat Q_{\bm \pi \circ \bm v}^i\left(s,\bm a\right)=\mathbb E_{\bm \pi \circ \bm v,p}\left(\sum_{k=0}^\infty \gamma^kr^i_{t+k+1}|s_t=s,\bm a_t= \bm a\right),
\end{equation}
where $\bm \pi\in \mathcal\prod^1\times\cdots\times\prod^N: \mathcal S \rightarrow\mathcal{A}^1\times\cdots\times\mathcal{A}^N$ is the joint policy, $\pi^i\in\prod^i: \mathcal S \rightarrow\mathcal{A}^i$ is the policy of agent $i$, $\prod^i$ is the policy space of agent $i$, $\bm v\mathop=\limits^\Delta \left(v^i,\cdots,v^M\right)$ denotes the joint adversarial perturbation and $\bm \pi \circ \bm v$ denotes the joint policy under the joint adversarial perturbation.


The value function and the action-value function can respectively be written as follows:
\begin{equation}
\widehat V_{{\bm{\pi }} \circ {{\bm{v}}_*}}^i\left( s \right) = \min_{v^i} \widehat V_{{\bm{\pi }} \circ\left( {v^i,\bm{v}^{-i}_*}\right)}^i\left( s \right),
\end{equation}
\begin{equation}
	\widehat Q_{{\bm{\pi }} \circ {{\bm{v}}_*}}^i\left( s, \bm{a} \right) = \min_{v^i} \widehat Q_{{\bm{\pi }} \circ\left( {v^i,\bm{v}_*^{-i}}\right)}^i\left( s,\bm{a} \right),
\end{equation}
where $\bm v_*\mathop=\limits^\Delta\left(v^1_*,\cdots,v^M_*\right)$ is the joint optimal adversarial perturbation, $v^i$ is an arbitrary valid adversarial perturbation and $\bm v_*^{-i}\mathop=\limits^\Delta\left(v_*^1,\cdots,v_*^{i-1},v_*^{i+1},\cdots,v_*^M\right)$. The goal of the joint optimal adversarial perturbation is to minimize the expected cumulative discount reward of every victim agent.

In \cite{zhou2022romfac}, the properties of SASG are discussed, including the existence and contraction of the joint optimal adversarial perturbation. And they point out that solving the joint optimal adversarial perturbation is equal to solving an SG which can be solved by MARL. Therefore, there is some research to solve this one via MARL \cite{han2022solution, li2023attacking, 9283830}. However, the victim agents are certain in the training process. If the victims are changing, MARL models have to retrain. We solve this problem in the next section.
\section{Robustness Testing on Critical Agent States}
In this section, we define agent State Adversarial Dec-POMDPs (SA-Dec-POMDPs) and propose a novel framework to solve it. 
\begin{figure}[t]
	\vskip 0.2in
	\begin{center}
	\centerline{\includegraphics[scale=0.8]{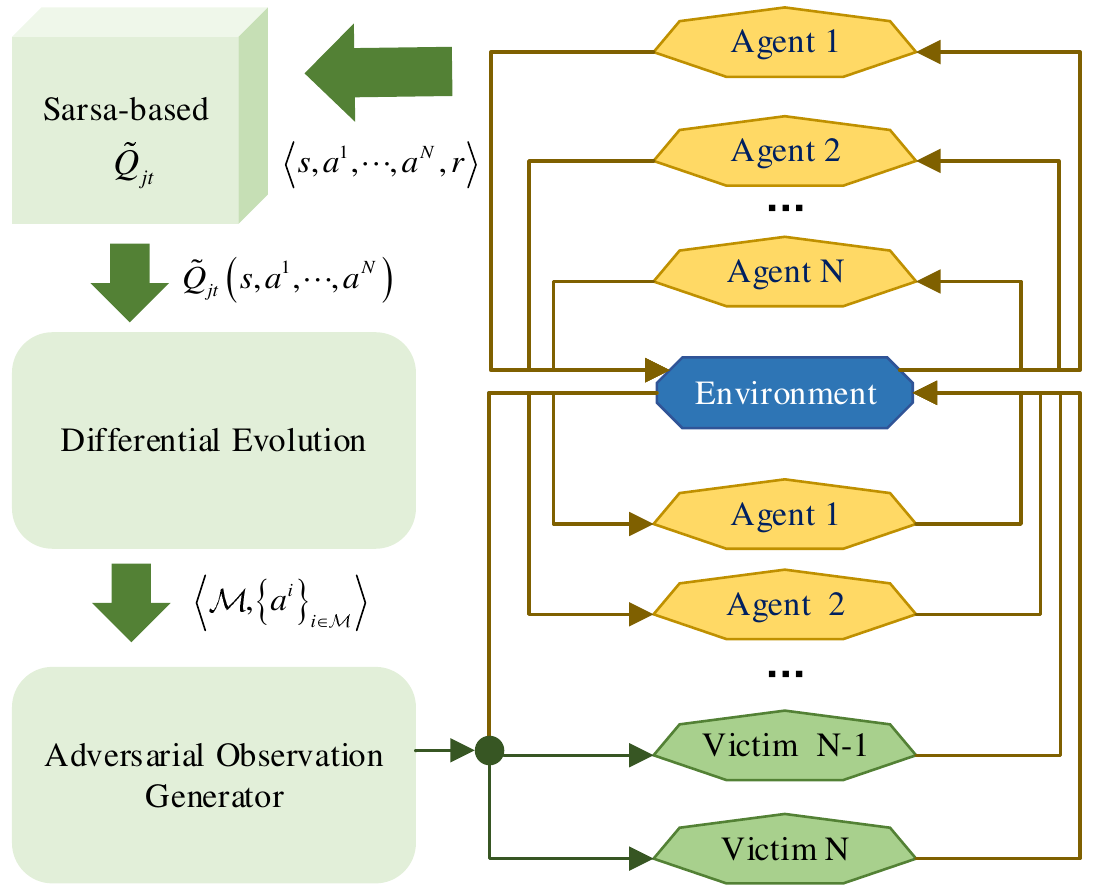} }
	\caption{The illustration of RTCA.} 
	\label{SA-SG} 
\end{center}
\vskip -0.2in
\end{figure}
\begin{definition}[SA-Dec-POMDPs]
An SA-Dec-POMDP is defined as a tuple $$\left<\mathcal{S}, \mathcal\{{A}^i\}_{i\in\mathcal{N}}, \{\mathcal{O}^i\}_{i\in\mathcal{N}}, \{\mathcal{B}^i\}_{i\in\mathcal{M}}, \mathcal{N}, \mathcal{M}, Z, r, p,  \gamma\right>$$ where $\mathcal{B}$ is the set of the adversarial observation of agent $i$. Assume that the adversarial perturbation $v^i(\tau^i)$ of agent $i$ is a deterministic function $v^i:\mathcal T^i \rightarrow \mathcal B^i$.  

\end{definition}
Similar to SASG, the adversary aims to minimize the expected cumulative discount reward of the team via manipulating the observation of the victim. The value function and the action-value function of the adversary can be written as follows:
\begin{equation}
\widehat V_{\pi^i  \circ {v}^i_*}^i\left( \tau^i \right) = \min_{v^i} \widehat V_{\pi^i \circ {v^i}}^i\left( \tau^i \right),
\end{equation}
\begin{equation}
	\widehat Q_{\pi^i \circ v^i_*}^i\left(\tau^i, a^i \right) = \min_{v^i} \widehat Q_{\pi \circ v^i}^i\left( \tau^i, a^i \right).
\end{equation}
Obviously, solving the optimal joint policy for the adversary in SA-Dec-POMDPs is equivalent to solving the optimal joint policy for the agent in Dec-POMDPs. The action space of the agent in Dec-POMDPs is the adversarial observation space of the adversary in SA-Dec-POMDPs. Given this, we can leverage CTDE-based MARL to address this problem. However, it should be noted that victim agents are known with certainty, including the number of victims $M$ and their respective indexes. When the value of $M$ changes, the SA-Dec-POMDP also changes, it is necessary to retrain the adversary agent via CTDE-based MARL.

Not only in SA-Dec-POMDPs but SG and SASG, the set of agents $\mathcal N$ is a certainty. And learning a MARL model to generate adversarial observations is challenging when the space of one $\mathcal{B}^M$ is large.
In this paper, we propose a novel robustness testing framework, i.e., solving of SA-Dec-POMDPs named RTCA as shown in Figure \ref{SA-SG} that can choose critical agents from the set of $\mathcal{N}$ as the set of victims $\mathcal{M}$ and provide guidance on the worst joint action of victims, followed by generating adversarial observations on them. Furthermore, our RTCA does not require any training, allowing for a variable set of victims $\mathcal M$ to be used. 

\subsection{Selection of critical agents and worst actions}
We present a method based on DE for selecting critical agents and worst joint actions in MARL. The goal of this method is to identify those agents and their actions that have the greatest influence on the overall performance of the system.

Motivated by CTDE, if the joint action-value function is known for the adversary, the objective of the adversary is as follows:
\begin{equation}
    v_{de}\left(\bm \tau\right) = \arg \mathop {\min }\limits_{\mathcal{M},\{a^i_{i\in \mathcal{M}}\}}Q_{jt}\left(\bm \tau, \bm a^M,\bm a^{-M}\right)
\end{equation}
where $\bm a^M$ and $\bm a^{-M}$ are the joint action of the victim and the other agents, respectively. We employ DE to accomplish this objective. The DE algorithm is a global optimization algorithm \cite{DE, BILAL2020103479, OPARA2019546} that the concept of population evolution for exploring the global optimal solution and is extensively applied to solve complex optimization problems. 
It maintains a population of candidate solutions, which are represented as vectors in a high-dimensional search space. During each iteration of the DE algorithm, new candidate solutions are generated by perturbing the vectors in the population and combining them using a simple arithmetic operator. The fitness of candidate solutions is then evaluated using the objective function, and the best ones are selected to form the next generation of the population.

The main advantage of the DE algorithm is the use of a differential mutation operator, which perturbs the candidate solutions in a way that is guided by the difference between two randomly selected vectors from the population. This operator allows the algorithm to explore the search space efficiently and to converge to a high-quality solution in a relatively short time.

In this paper, we encode the critical agent and the corresponding worst joint action into a candidate solution. One candidate solution contains the actions of $M$ victims and that is a tuple of $2M$ elements: the indexes of critical agents and their joint worst actions ${a^i}_{i\in\mathcal{M}}$. We use the same setting with \cite{8601309}. At the start of the algorithm, a population of $400$ candidate solutions $D$ is generated. During every iteration, an additional $400$ candidate solutions are generated utilizing the following standard DE formula:
\begin{equation}\label{d}
\begin{aligned}
    D_{j}\left(g+1\right) = D_{d_1}&\left(g\right) + F\left(D_{d_2}\left(g\right)-D_{d_3}\left(g\right)\right)\\
    &d_1 \ne d_2 \ne d_3,
\end{aligned}
\end{equation}
where $g$ is the current generation index, and each element $D_{j}$ of the candidate solution is combined with three randomly chosen elements using a scaling parameter $F$ set to $0.5$. The indices of the randomly chosen elements are represented by $d_1$, $d_2$, and $d_3$. The whole process is presented in Algorithm \ref{de}.
\begin{algorithm}[t]
\caption{Selection of critical agents and worst actions}
\label{de}
\KwIn{Objective function $Q_{jt}$, the joint action-observation histories $\bm \tau$, the joint policy of agents $\bm \pi$, the value $Q_{jt}\left(\bm \tau, \bm a\right)$, population size $400$, scaling factor $F=0.5$, crossover rate $CR$, maximum number of iterations $T$}
\KwOut{$D_{best}\in D \mathop=\limits^\Delta\{\mathcal M, \{\hat a^i\}_{i\in \mathcal M}\}$}
Initialize population $D$ with $400$ random solutions; and 
Evaluate the fitness of each solution in the population based on $Q_{jt}$;
\While{$t=1,2,\cdots,T$}{
\For{each solution $x_j \in D$}{
Randomly select three distinct solutions $D_{d_1}$, $D_{d_2}$ and $D_{d_3}$ from $D$;\\
Generate a mutant vector $D_j$ according to (\ref{d});\\
Generate a trial vector $u_j$ by performing a binary crossover between $x_j$ and $D_j$ with crossover rate $CR$;\\
Generate joint action of the victim agent $\bm a^M_u$ and $\bm a^M_D$ based on $u_j$ and $D_j$, respectively.\\
Evaluate the fitness of $u_j$;\\
\If {$Q_{jt}(\bm \tau, \bm a^M_u, \bm a^{-M}) < Q_{jt}(\bm \tau, \bm a^M_D, \bm a^{-M})$}{
Replace $D_j$ with $u_j$ in the population $D$;
}
}
$t=t+1$;
}
Select and return the best solution found, $D_{best}$, from $P$.
\end{algorithm}

By doing so, we can identify which agents are most critical to the team performance, and which actions on them are most likely to lead to failure or sub-optimal outcomes. The advantages of using DE to compute the index of critical agents and their joint worst actions are as follows:
\begin{itemize}
    \item Less information. The joint action-value function is not necessarily differentiability. In fact, the joint action-value network of QMIX is non-differentiability making it difficult for the adversary to estimate the team reward. Furthermore, we consider the case of a discrete action space and the situation where a subset of agents becomes the victim. Even if the joint action-value network is differentiable, such as in VDN, gradient-based attack methods cannot be used to solve the worst-case joint action. DE is ideal for addressing this type of problem.
    \item More flexibility. It is not necessary to know the structure of the joint action-value function; only its inputs and outputs are required. Therefore, networks trained in any way can be attacked. In addition, the number of victims $M$ only needs to remain constant during the current time step and can be changed arbitrarily in the future without the need for retraining.
\end{itemize}

\subsection{Sarsa-based joint action-value function}

In Section 4.1, we assume that the joint action-value function is known to the adversary. However, in reality, this component is not being deployed. We cannot directly compute $Q_{jt}$, and more importantly, the training of $Q_{jt}$ in CTDE is based on the assumption that all agents make the optimal decision. As a result, the joint action-value function in VDN or QMIX may not evaluate the team reward. To solve this problem, we train a joint action-value network based on Sarsa \cite{rummery1994line} during the execution process. 

Since the $\bm \pi$ is fixed in the execution process, we use $\epsilon$-greedy to explore all situations possible, not just the good trajectories. We construct the joint action-value function $\Tilde Q_{jt}$ as a neural network with parameters $\theta$. Its input is the environment state $s$ and the joint action of agent $\bm a$. Learning object of $\Tilde Q_{jt}$ is to minimize:
\begin{equation}\label{ljt}
    L_{jt} \left(\theta\right) = \sum_{k=1}^{BS}\left[R_k+\gamma\Tilde{Q}_{jt}\left(s'_k,\bm a'_k\right)-\Tilde{Q}_{jt}\left(s_k,\bm a_k\right)\right],
\end{equation}
where $BS$ is the batch size. The training process is shown in Algorithm \ref{cq}.
\begin{algorithm}[t]
\caption{Sarsa-based joint action-value function}
\label{cq}
Initialize $\Tilde Q_{jt}^{\theta}$ and replay buffer $\mathcal{R}$;\\
\For{$t = 1,2,\dots,T$}{
Get the current state of the environment $s$ and the observation of each agent $\left[o^1,\cdots,o^N\right]$;\\
Take the $\epsilon$-greedy joint action $\bm a$ according to $\left[Q_1\left(\tau^1,a^1\right), \cdots, Q_N\left(\tau^N,a^N\right)\right]$ and get the next state $s'$ and observation $\left[o'^1,\cdots,o'^N\right]$;\\
Put $\left<s, \bm a, r, s'\right>$ in $\mathcal{R}$;\\
\If{$|\mathcal{R}| \ge BS$}{
Sample $BS$ transitions from replay buffer $\mathcal{R}$ $\{\left<s_k, \bm a_k, r_k, s'_k\right>\}_{k=1,2,\cdots, BS}$;\\
Update the parameters $\theta$ via minimizing (\ref{ljt}). 
}
}

\end{algorithm}
\begin{algorithm}[t]
\caption{RTCA}
\label{RTCAa}
Initialize the environment.\\
\For{$t = 1,2,\dots,T$}{
Get the current state of the environment $s$ and the observation of each agent $\left[o^1,\cdots,o^N\right]$;\\
Generate the victim index and the worst action according to Algorithms \ref{de} and \ref{cq};\\
Generate the adversarial state according to (\ref{fgsm});\\
Take action based on the clean observation for the normal agent and based on the adversarial observation for the victim agent.
}

\end{algorithm}
\subsection{Generation of adversarial observations}
After obtaining the index of the critical agent and the worst joint action, we aim to generate adversarial perturbation using these to induce the victim to take the worst actions. 

In the classification tasks, there are many methods of targeted attacks\cite{goodfellow2014explaining, Li_2020_CVPR, madry2018towards}, which add the perturbation to the sample to make the neural network output the targeted label. In our setting, this type of method can be used to generate adversarial perturbations based on the policy of the victim agents, misleading them to output the targeted action. This process can be achieved by optimizing the following objective function,
\begin{equation} \label{advloss}
\begin{aligned}
	L_{sa} \left(\theta^i, \hat{o}^i\right)=\min \limits_{\hat{o}^i \in \mathcal{B}^i} \bigg\{-\sum_{a^j\in\mathcal{A}^i} \left[\hat a^i = a^j\right] \log\left({\pi_{\theta^i}}\left(a^j|\hat o^i\right)\right) \\
 +\sum_{a^j\in\mathcal{A}^i} \left[z^i = a^j\right] \log\left({\pi_{\theta^i}}\left(a^j|\hat o^i\right)\right)\bigg\},
 \end{aligned}
\end{equation}
where $z^i$ is the action of victim agent $i$ based on the clean observation, $\theta^i$ is the parameter of the policy network of victim agent $i$, $\hat a^i$ is the target action of victim agent $i$ and $\hat o^i$ is the perturbation observation of agent $i$. This optimization makes the victim take the action which is closer to the target action and farther to the true one. We use Fast Gradient Sign Method (FGSM) \cite{goodfellow2014explaining} to address this optimization problem. FGSM uses the one-step calculation of gradient descent as follows:
\begin{equation} \label{fgsm}
	\hat o^i=\mbox{clip}\left(\hat o^i +  \alpha \mbox{sgn}\left(\nabla_{\hat o^i}  L_{sa}\left(\theta^i,\hat o^i, \hat a^i\right) \right)\right)
\end{equation}
where $\alpha$ is the step size. 
If the effective value range of $o$ is $[m,n]$, then
\begin{equation}
\mbox {clip}\left( o \right) = \left\{ \begin{array}{l}
o,o \in \left[ {m,n} \right]\\
m,o < m\\
n,o > n
\end{array} \right..
\end{equation}
Our RTCA is presented in Algorithm \ref{RTCAa}.

\begin{figure*}[t]
	\vskip 0.2in
	\begin{center}
	\centerline{\includegraphics[scale=0.48]{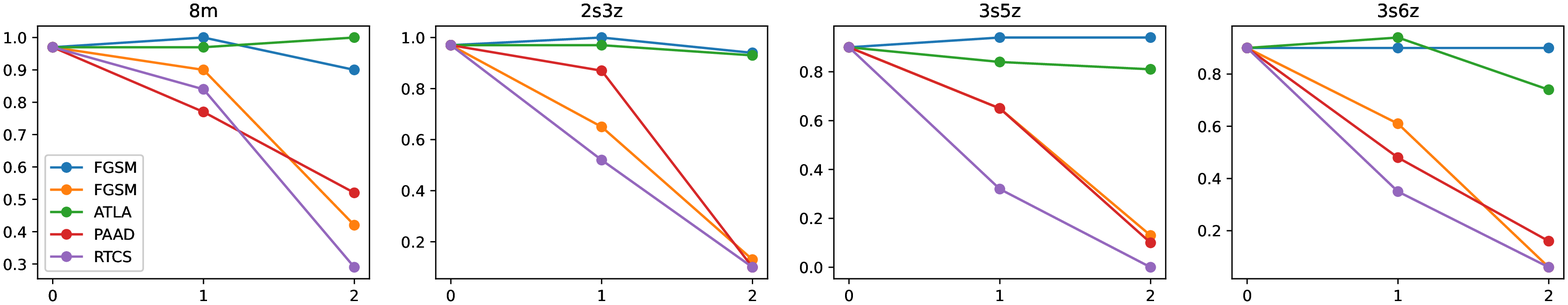} }
	\caption{The illustration of robustness test for QMIX. The horizontal coordinate indicates the number of victims and the vertical coordinate indicates the winning rate.} 
	\label{re_qmix} 
\end{center}
\vskip -0.2in
\end{figure*}

\begin{figure*}[t]
	\vskip 0.2in
	\begin{center}
	\centerline{\includegraphics[scale=0.48]{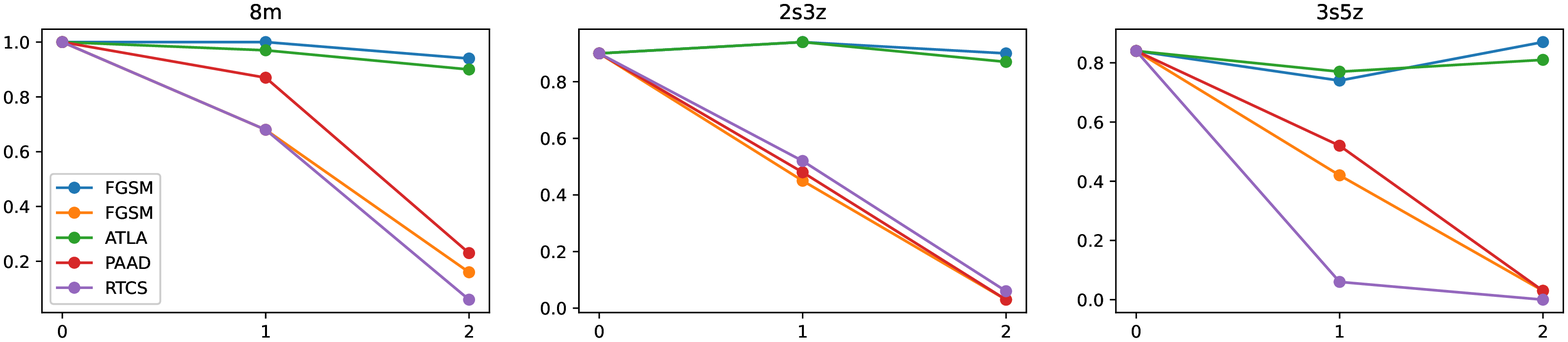} }
	\caption{The illustration of robustness test for VDN. The horizontal coordinate indicates the number of victims and the vertical coordinate indicates the winning rate.} 
	\label{re_vdn} 
\end{center}
\vskip -0.2in
\end{figure*}
\begin{table*}[t]
\centering
\caption{The performance of QMIX and VDN in the number of victims is 0, 1, and 2 under the random noise. WR is the winning rate. $|M|$ is the number of victims.}
\label{rand}
\begin{tabular}{cccccccc}
\toprule[0.8pt]
\multirow{2}{*}{Envs} & Number of victims & \multicolumn{2}{c}{0}                     & \multicolumn{2}{c}{1}                     & \multicolumn{2}{c}{2}                     \\ \cline{2-8} 
                          & Victim            & \multicolumn{1}{c}{WR}  & Reward         & \multicolumn{1}{c}{WR}  & Reward         & \multicolumn{1}{c}{WR}  & Reward         \\ \toprule[0.6pt]
\multirow{2}{*}{8m}       & QMIX              & \multicolumn{1}{c}{0.97} & 19.74$\pm$1.45 & \multicolumn{1}{c}{1.00} & 20.00          & \multicolumn{1}{c}{0.90} & 19.31$\pm$2.10 \\ 
                          & VDN               & \multicolumn{1}{c}{1.00} & 20.00          & \multicolumn{1}{c}{1.00} & 20.00          & \multicolumn{1}{c}{0.94} & 19.48$\pm$1.99 \\ \hline
\multirow{2}{*}{2s3z}     & QMIX              & \multicolumn{1}{c}{0.97} & 19.73$\pm$1.47 & \multicolumn{1}{c}{1.00} & 20.00          & \multicolumn{1}{c}{0.94} & 19.72$\pm$1.08 \\ 
                          & VDN               & \multicolumn{1}{c}{0.90} & 19.58$\pm$1.41 & \multicolumn{1}{c}{0.94} & 19.64$\pm$1.43 & \multicolumn{1}{c}{0.90} & 19.55$\pm$1.43 \\ \hline
\multirow{2}{*}{3s5z}     & QMIX              & \multicolumn{1}{c}{0.90} & 19.52$\pm$1.52 & \multicolumn{1}{c}{0.94} & 19.73$\pm$1.09 & \multicolumn{1}{c}{0.94} & 19.69$\pm$1.28 \\ 
                          & VDN               & \multicolumn{1}{c}{0.84} & 19.22$\pm$1.94 & \multicolumn{1}{c}{0.74} & 18.90$\pm$2.09 & \multicolumn{1}{c}{0.87} & 19.44$\pm$1.61 \\ \hline
3s6z                      & QMIX              & \multicolumn{1}{c}{0.90} & 19.48$\pm$1.64 & \multicolumn{1}{c}{0.90} & 19.45$\pm$1.70 & \multicolumn{1}{c}{0.90} & 19.37$\pm$1.94 \\  
\toprule[0.8pt]
\end{tabular}
\end{table*}

\begin{table*}[t]
	\caption{The results of robustness testing for QMIX and VDN. WR is the winning rate. $M$ is the number of victims. The smaller the WR and reward, the better. Bold scores represent the best performance.}
        \label{result}
	\centering
	\begin{tabular}{ccccccccccc}
		\toprule[0.8pt]
		\multirow{2}{*}{Envs} & Adversary             & \multicolumn{2}{c}{FGSM} & \multicolumn{2}{c}{ATLA} & \multicolumn{2}{c}{PAAD} & \multicolumn{2}{c}{RTCA} & \multirow{2}{*}{$M$} \\ \cline{2-10}
		& Victim                & WR     & Reward          & WR     & Reward          & WR     & Reward          & WR     & Reward          &                        \\  
		\toprule[0.6pt]
		\multirow{4}{*}{8m}   & QMIX                  & 0.90   & 19.25$\pm$2.30  & 0.97   & 19.79$\pm$1.18  & \textbf{0.77}   & 18.18$\pm$3.45  & 0.84   & 18.86$\pm$2.61  & \multirow{2}{*}{1}     \\
		& VDN                   & \textbf{0.68}   & 17.24$\pm$4.04  & 0.97   & 19.75$\pm$1.34  & 0.87   & 19.04$\pm$2.49  & \textbf{0.68}   & 17.36$\pm$3.87  &                        \\ \cline{2-11} 
		& QMIX                  & 0.42   & 15.57$\pm$3.82  & 1.00   & 20.00           & 0.52   & 16.05$\pm$4.15  & \textbf{0.29}   & 14.20$\pm$3.83  & \multirow{2}{*}{2}     \\
		& VDN                   & 0.16   & 12.30$\pm$3.62  & 0.90   & 19.31$\pm$2.12  & 0.23   & 13.15$\pm$3.86  & \textbf{0.06}   & 10.78$\pm$2.79  &                        \\ \hline
		\multirow{4}{*}{2s3z} & QMIX                  & 0.65   & 17.87$\pm$3.13  & 0.97   & 19.87$\pm$0.79  & 0.87   & 19.08$\pm$2.41  & \textbf{0.52}   & 16.93$\pm$3.43  & \multirow{2}{*}{1}     \\
		& VDN                   & \textbf{0.45}   & 16.64$\pm$3.34  & 0.94   & 19.67$\pm$1.44  & 0.48   & 16.89$\pm$3.29  & 0.52   & 16.97$\pm$3.37  &                        \\ \cline{2-11}
		& QMIX                  & 0.13   & 13.29$\pm$2.96  & 0.93   & 19.52$\pm$1.85  & \textbf{0.10}   & 14.27$\pm$2.32  & \textbf{0.10}   & 13.75$\pm$2.73  & \multirow{2}{*}{2}     \\
		& VDN                   & \textbf{0.03}   & 11.81$\pm$1.98  & 0.87   & 19.25$\pm$2.02  & \textbf{0.03}   & 12.33$\pm$2.29  & 0.06   & 12.49$\pm$2.48  &                        \\ \hline
		\multirow{4}{*}{3s5z} & QMIX                  & 0.65   & 17.92$\pm$3.03  & 0.84   & 19.26$\pm$1.79  & 0.65   & 18.19$\pm$2.63  & \textbf{0.32}   & 16.35$\pm$2.91  & \multirow{2}{*}{1}     \\
		& VDN                   & 0.42   & 16.48$\pm$3.39  & 0.77   & 19.21$\pm$1.76  & 0.52   & 17.49$\pm$3.12  & \textbf{0.06}   & 14.25$\pm$2.69  &                        \\ \cline{2-11}
		& QMIX                  & 0.13   & 13.83$\pm$2.83  & 0.81   & 19.13$\pm$1.85  & 0.10   & 13.23$\pm$2.72  & \textbf{0.00}   & 12.70$\pm$1.50  & \multirow{2}{*}{2}     \\
		& VDN                   & 0.03   & 10.64$\pm$2.69  & 0.81   & 19.49$\pm$1.32  & 0.03   & 11.95$\pm$3.01  & \textbf{0.00}   & 9.71$\pm$1.51   &                        \\ \hline
		\multirow{2}{*}{3s6z} & \multirow{2}{*}{QMIX} & 0.61   & 18.11$\pm$2.66  & 0.94   & 19.71$\pm$1.12  & 0.48   & 17.15$\pm$3.10  & \textbf{0.35}   & 16.48$\pm$3.02  & 1   \\ \cline{3-11} 
		&                       & 0.06   & 13.15$\pm$2.56  & 0.74   & 18.97$\pm$2.02  & 0.16   & 14.08$\pm$3.05  & \textbf{0.06}   & 13.40$\pm$2.58  & 2                     \\ 
		\toprule[0.8pt]
	\end{tabular}
\end{table*}

\section{Experiment Results and Analysis}
In this section, we demonstrate the outstanding performance of RTCA in generating adversarial observation from the perspective of the number of victims and decreasing the team reward.

\subsection{Experiment settings}

\subsubsection{Environment settings}
We evaluate our robustness testing framework on the StarCraft Multi-Agent Challenge (SMAC) \cite{10.5555/3306127.3332052}. SMAC is a distributed real-time strategy game widely used to evaluate the performance of RL. The main task is to defeat the opponent through cooperation among agents. We conduct experiments on four maps containing 8 Marines (8m), 2 Stalkers \& 3 Zealots (2s3z), 3 Stalkers \& 5 Zealots (3s5z), and 3 Stalkers \& 6 Zealots (3s6z).  
\begin{itemize}
    \item Observation space: At each time step, the agents obtain information about their field of view, including the following information about their opponent and teammates: distance, relative $x, y$, health, shield, and unit type. The state of the environment contains information about all units on the map. Specifically, the coordinates of all agents relative to the center of the map, as well as the elements present in their observations are contained. The global state is only used in the centralized training process. During the testing process, we only attack the observations of the agents.
    \item Action space: In the SMAC environment, each agent has access to four possible actions: move, attacking opponent, stop, and no-op. Movement is restricted to four cardinal directions: north, south, east, or west. The attack can only be performed if the enemy unit is within the shooting range. The stop is to do nothing and the no-op can take when the agent is dead.
    \item Reward: The overall goal is to increase the win rate of the agents as much as possible. SMAC uses shaped reward by default, and at each time step, the agents receive a reward based on the hit point damage. In addition, the agent receives an additional reward for each opponent it defeats and for winning by defeating all opponents. All of these rewards are normalized to ensure that the maximum cumulative reward earned in an episode is $20$.
\end{itemize}

\subsubsection{Benchmark methods}
The victim agent is trained via VDN and QMIX with two million steps. We compare RTCA with the following state-of-art robustness testing method. 
\begin{itemize}
    \item Random noise. The adversary adds some random perturbations to the observation of the victim agent. We use uniform distribution noise as a way to interfere with the decision-making of the victim.
    \item FGSM \cite{goodfellow2014explaining}. The adversary generates the adversarial perturbation according to the gradient on the decision-making network of the victim. It does not consider the effect of the perturbation on team cumulative reward, but only the disruption of individual policy, i.e., making the victim do actions that are inconsistent with those in the clean observation.
    \item ATLA \cite{zhang2021robust}. The adversary is trained via PPO \cite{schulman2017proximal} in the single-agent scenario. We use MAPPO \cite{NEURIPS2022_9c1535a0} as adversaries in the multi-agent scenario. For MAPPO agents, the input is the clean observation, the action is the adversarial observation, and the reward is the negative of victimization teams. In the training process, we set all agents in the environment are victims.
    \item PAAD \cite{sun2022who}. The adversary is trained in two steps including the director giving advice for the worst joint action and the actor generating the adversarial observation based on this action. The director is a PPO agent in the single-agent scenario. We use the same method (i.e., VDN or QMIX) with victims to train MARL directors and set all agents in the environment are victims.
\end{itemize}

\subsubsection{Perturbation settings}
The perturbation is set as a $\ell_{\infty}$ with a range of $0.1$. In our RTCA, the victim set is changed at each time step. For fairness, it is randomly changed in the benchmark methods. We conduct $32$ episodes and use winning rates and average team cumulative rewards to evaluate the performance of those attack methods.

\subsection{Results and discussions}

The experiment results are shown in Tables \ref{rand} and \ref{result}. 
Figures \ref{re_qmix}  and \ref{re_vdn} provide a more intuitive presentation of the experimental results. The robustness of VDN is not evaluated in the 3s6z scenario as its performance is lacking. 

In the 8m scenario, the agents are homogeneous and play the same role in the team. The purpose of IGM in MARL is to solve the problem of lazy agents and make the contribution of each agent to the team as equal as possible. However, from our experimental results, it can be seen that the overall effect of selecting critical agents as victims is better than random selection. There might exist more than one critical agent because each on the team in the 8m scenario performs the same job, and there is an opportunity that the crucial agent can be chosen at random. As a result, when attacking just one agent, the PAAD outperforms the RTCA in terms of QMIX. However, the average team cumulative reward based on PAAD has a high variance, indicating that there is some randomness in which the agent is chosen as a critical one. Our strategy demonstrates superior performance in attacking QMIX in scenarios with heterogeneous agents. For the robustness testing of VDN, RTCA only performs poorly in the 2s3z scenario but is similar to PAAD and FGSM. The performance of the model trained by VDN in the 2s3z scenario is poor which causes the $\tilde Q_{jt}$ learned to be worse, making it difficult for the adversary to accurately compute the worst joint action.

The attack results of random noise are very poor and even enhance the winning rate of the victim, which indicates that some noise has a positive effect on the decision-making of MARL agents. This is not appropriate for finding the robustness fault of MARL. Likewise, the ATLA method has poor results, similar to those of random noise. We analyze the reason for this and believe that it is because the joint observation space of agents grows exponentially with the number of agents in a multi-agent task. This is very difficult for MAPPO agents to learn the relationship between clean observations and their joint actions (i.e., joint adversarial observations). Thus, its effectiveness is poor. Overall, we can conclude that:
\begin{itemize}
    \item Random noise does not consider the policy of the victim, therefore it is a weak attack method.
    \item FGSM only considers the policy of the victim, not the team cooperation. Bad actions by the victim do not necessarily lead to the worst for the team.
    \item ATLA has a large action space making it difficult to generate the adversarial observation.
    \item PAAD considers the effect of victims on the team, but the victim must be certain. 
    \item RTCA is the most suitable for robustness testing of MARL due to its flexibility and strong attack. Furthermore, this kind of attack is more stealthy because the victim agent changes at every time step.
\end{itemize}
\begin{table}
\caption{Ablation study. The robustness of QMIX and VDN is evaluated based on their action-value functions. $M$ is the number of victims}
\begin{center}
\label{abl}
\begin{tabular}{ccccccc}
\toprule[0.6pt]
\multirow{2}{*}{Envs} & $Q_{jt}$             & \multicolumn{2}{c}{VDN/QMIX}            & \multicolumn{2}{c}{$\tilde Q_{jt}$}                  & \multirow{2}{*}{$M$} \\ \cline{2-6}
                          & Victim                & \multicolumn{1}{l}{WR}  & Reward         & \multicolumn{1}{c}{WR}  & Reward         &                        \\ \hline
\multirow{4}{*}{8m}       & QMIX                  & \multicolumn{1}{l}{0.74} & 18.15$\pm$3.14 & \multicolumn{1}{c}{0.84} & 18.86$\pm$2.61 & \multirow{2}{*}{1}     \\ 
                          & VDN                   & \multicolumn{1}{l}{0.84} & 18.60$\pm$3.19 & \multicolumn{1}{c}{0.68} & 17.36$\pm$3.87 &                        \\ \cline{2-7} 
                          & QMIX                  & \multicolumn{1}{l}{0.35} & 14.49$\pm$4.21 & \multicolumn{1}{c}{0.29} & 14.20$\pm$3.83 & \multirow{2}{*}{2}     \\ 
                          & VDN                   & \multicolumn{1}{l}{0.16} & 12.23$\pm$3.65 & \multicolumn{1}{c}{0.06} & 10.78$\pm$2.79 &                        \\ \hline
\multirow{4}{*}{2s3z}     & QMIX                  & \multicolumn{1}{l}{0.52} & 17.02$\pm$3.26 & \multicolumn{1}{c}{0.52} & 16.93$\pm$3.43 & \multirow{2}{*}{1}     \\ 
                          & VDN                   & \multicolumn{1}{l}{0.87} & 19.41$\pm$1.54 & \multicolumn{1}{c}{0.52} & 16.97$\pm$3.37 &                        \\ \cline{2-7} 
                          & QMIX                  & \multicolumn{1}{l}{0.10} & 13.44$\pm$2.53 & \multicolumn{1}{c}{0.10} & 13.75$\pm$2.73 & \multirow{2}{*}{2}     \\ 
                          & VDN                   & \multicolumn{1}{l}{0.39} & 15.90$\pm$3.73 & \multicolumn{1}{c}{0.06} & 12.49$\pm$2.48 &                        \\ \hline
\multirow{4}{*}{3s5z}     & QMIX                  & \multicolumn{1}{l}{0.48} & 17.08$\pm$3.11 & \multicolumn{1}{c}{0.32} & 16.35$\pm$2.91 & \multirow{2}{*}{1}     \\ 
                          & VDN                   & \multicolumn{1}{l}{0.42} & 16.69$\pm$3.27 & \multicolumn{1}{c}{0.06} & 14.25$\pm$2.69 &                        \\ \cline{2-7} 
                          & QMIX                  & \multicolumn{1}{l}{0.03} & 13.14$\pm$2.08 & \multicolumn{1}{c}{0.00} & 12.70$\pm$1.50 & \multirow{2}{*}{2}     \\ 
                          & VDN                   & \multicolumn{1}{l}{0.00} & 11.70$\pm$2.17 & \multicolumn{1}{c}{0.00} & 9.71$\pm$1.51  &                        \\ \hline
\multirow{2}{*}{3s6z}     & \multirow{2}{*}{QMIX} & \multicolumn{1}{l}{0.39} & 17.06$\pm$2.75 & \multicolumn{1}{c}{0.35} & 16.48$\pm$3.02 & 1                      \\ \cline{3-7} 
                          &                       & \multicolumn{1}{l}{0.00} & 13.45$\pm$1.97 & \multicolumn{1}{c}{0.06} & 13.40$\pm$2.58 & 2                      \\ \hline
                          \toprule[0.6pt]
\end{tabular}
\end{center}
\end{table}
\subsection{Ablation study}
We use the value decomposition network and mixing network in VDN and QMIX, respectively, to indicate the impact of the action-value function $\tilde Q_{jt}$. The results are demonstrated in Table \ref{abl}. The joint action-value function of QMIX more accurately represents the quality of the joint policy than VDN. Therefore, critical agents can be chosen as victims and their worst joint actions can be calculated using $Q_{jt}$ of QMIX.  According to the experimental results, it can be seen that using $\tilde Q_{jt}$ achieves results comparable to QMIX, and even better in complex scenarios (3s5z and 3s6z). On the other hand, because $Q_{jt}$ of VDN cannot accurately evaluate the quality of the joint action well,  it is not advisable to employ it as the objective function of DE.

\subsection{Transferability of $\tilde Q_{jt}$}
The purpose of this section is to evaluate the transferability of the $\tilde Q_{jt}$. In this context, transferability is defined as the ability of the $\tilde Q_{jt}$ trained using the data sampled by one MARL to transfer the experience to another and successfully attack it.
The results present in Table \ref{trans} indicate that the $\tilde Q_{jt}$ learned via QMIX policy sampling can be successfully attacked against VDN agents in a majority of the cases considered. Correspondingly, the $\tilde Q_{jt}$ trained with VDN policy sampling demonstrated high success rates in successfully attacking QMIX agents in most of the cases. Such observations hold significant implications for improving the applicability of adversarial attacks.
In a word, the Sarsa-based $\tilde Q_{jt}$ exhibits advantageous transferability qualities and possesses considerable prospects for usage as a black-box attack.
\begin{table}
\centering
\caption{Transferability of $\tilde Q_{jt}$. $M$ is the number of victims.}
\label{trans}
\begin{tabular}{ccccccc}
\toprule[0.6pt]
\multirow{2}{*}{Envs} & $\tilde Q_{jt}$ & \multicolumn{2}{c}{QMIX}                  & \multicolumn{2}{c}{VDN}                   & \multirow{2}{*}{$M$} \\ \cline{2-6}
                      & Victim        & \multicolumn{1}{c}{WR}  & Reward         & \multicolumn{1}{c}{WR}  & Reward         &                        \\ \hline
\multirow{4}{*}{8m}   & QMIX          & \multicolumn{1}{c}{0.84} & 18.86$\pm$2.61 & \multicolumn{1}{c}{0.81} & 18.56$\pm$2.96 & \multirow{2}{*}{1}     \\ 
                      & VDN           & \multicolumn{1}{c}{0.29} & 13.83$\pm$4.11 & \multicolumn{1}{c}{0.68} & 17.36$\pm$3.87 &                        \\ \cline{2-7} 
                      & QMIX          & \multicolumn{1}{c}{0.29} & 14.20$\pm$3.83 & \multicolumn{1}{c}{0.48} & 15.96$\pm$3.98 & \multirow{2}{*}{2}     \\ 
                      & VDN           & \multicolumn{1}{c}{0.10}  & 11.63$\pm$2.95 & \multicolumn{1}{c}{0.06} & 10.78$\pm$2.79 &                        \\ \hline
\multirow{4}{*}{2s3z} & QMIX          & \multicolumn{1}{c}{0.52} & 16.93$\pm$3.43 & \multicolumn{1}{c}{0.55} & 17.11$\pm$3.29 & \multirow{2}{*}{1}     \\ 
                      & VDN           & \multicolumn{1}{c}{0.48} & 16.25$\pm$3.81 & \multicolumn{1}{c}{0.52} & 16.97$\pm$3.37 &                        \\ \cline{2-7} 
                      & QMIX          & \multicolumn{1}{c}{0.10} & 13.75$\pm$2.73 & \multicolumn{1}{c}{0.03} & 13.37$\pm$1.98 & \multirow{2}{*}{2}     \\ 
                      & VDN           & \multicolumn{1}{c}{0.00}    & 12.06$\pm$1.64 & \multicolumn{1}{c}{0.06} & 12.49$\pm$2.48 &                        \\ \hline
\multirow{4}{*}{3s5z} & QMIX          & \multicolumn{1}{c}{0.32} & 16.35$\pm$2.91 & \multicolumn{1}{c}{0.48} & 16.95$\pm$3.23 & \multirow{2}{*}{1}     \\ 
                      & VDN           & \multicolumn{1}{c}{0.13} & 14.05$\pm$3.02 & \multicolumn{1}{c}{0.06} & 14.25$\pm$2.69 &                        \\ \cline{2-7} 
                      & QMIX          & \multicolumn{1}{c}{0.00} & 12.70$\pm$1.50 & \multicolumn{1}{c}{0.03} & 13.52$\pm$1.91 & \multirow{2}{*}{2}     \\ 
                      & VDN           & \multicolumn{1}{c}{0.00}    & 10.00$\pm$1.65 & \multicolumn{1}{c}{0.00} & 9.71$\pm$1.51  &                        \\ 
                      \toprule[0.6pt]
\end{tabular}
\end{table}

\section{Related Work}
\subsection{Adversarial attacks on SARL} 
An extensive body of research has been conducted on methods related to generating adversarial examples in classification tasks. Furthermore, recent studies have emerged that explore adversarial attacks in the context of SARL.
Based on a survey~\cite{9536399}, adversarial attacks on SARL can be classified into four distinct categories including perturbations to the state space, the reward function, the action space, and the model space.
Huang et~al.~\cite{huang2017adversarial} utilize FGSM for creating adversarial examples of agent input states. Their findings illustrate the efficacy of adversarial attacks for the model trained by RL.
Pattanaik et~al.~\cite{10.5555/3237383.3238064} propose three types of methods containing random noise, gradient-based, and stochastic gradient decrease.
The sample efficient model-based adversarial attack is introduced by Weng et~al.~\cite{Weng2020Toward}. To achieve this, they propose a two-step attack framework including the learning for the dynamic environment model and the generation of the adversarial state based on the environment model.  
Huang et~al~\cite{NEURIPS2020_f0eb6568} propose the State-Adversarial Markov Decision Process (SA-MDP) which indicates the optimal adversary exists. Besides, they improve the gradient-based attack in \cite{10.5555/3237383.3238064} and propose a robust Sarsa attack. They use Sarsa to learn the critic network in continuous action space while we use Sarsa to train the joint action-value network in a discrete one. \cite{zhang2021robust} and \cite{sun2022who} are introduced in Section 4.

\subsection{Adversarial attacks on MARL} 
There are few studies on adversarial attacks in MARL. 
Lin et~al. \cite{9283830} propose the method of generating adversarial states for MARL, they use a two-step attack similar to \cite{sun2022who} which reduces the team reward by perturbing the state of only a fixed agent. 
Pham et~al.~\cite{pham2022evaluating} extend \cite{Weng2020Toward} to multi-agent setting. 
Guo et~al.~\cite{Guo_2022_CVPR} propose a comprehensive robustness testing framework named MARLSafe from three aspects: state, action, and reward. 
In these methods, the set of victim agents is fixed, while in RTCA, the set of victim agents is variable.

\section{Conclusion}
In this paper, we present a robust testing framework for a model trained by the state-of-the-art MARL. In our framework, based on DE, the critical agents are selected as victims and their worst joint actions are advised. Moreover, in order to evaluate the team cooperation policy, we present a Sarsa-based method to learn the joint action-value function in MARL. When the indexes of victims and their joint worst actions are known, we use the target attack method to generate the adversarial observation on them. The results clearly indicate the superiority of our RTCA over the existing ones. Our next work tends to apply RTCA to continuous action spaces such as MADDPG~\cite{NIPS2017_68a97503} and MAAC \cite{pmlr-v97-iqbal19a} to test their robustness against observation perturbations of critical agents.
\bibliographystyle{ieeetr}
\bibliography{ecai}

\begin{thebibliography}{10}

\bibitem{9103316}
T.~Wu, P.~Zhou, K.~Liu, Y.~Yuan, X.~Wang, H.~Huang, and D.~O. Wu, ``Multi-agent
  deep reinforcement learning for urban traffic light control in vehicular
  networks,'' {\em IEEE Transactions on Vehicular Technology}, vol.~69, no.~8,
  pp.~8243--8256, 2020.

\bibitem{9993797}
H.~Shi, G.~Liu, K.~Zhang, Z.~Zhou, and J.~Wang, ``Marl sim2real transfer:
  Merging physical reality with digital virtuality in metaverse,'' {\em IEEE
  Transactions on Systems, Man, and Cybernetics: Systems}, vol.~53, no.~4,
  pp.~2107--2117, 2023.

\bibitem{9209079}
L.~Wang, K.~Wang, C.~Pan, W.~Xu, N.~Aslam, and L.~Hanzo, ``Multi-agent deep
  reinforcement learning-based trajectory planning for multi-uav assisted
  mobile edge computing,'' {\em IEEE Transactions on Cognitive Communications
  and Networking}, vol.~7, no.~1, pp.~73--84, 2021.

\bibitem{vdn}
P.~Sunehag, G.~Lever, A.~Gruslys, W.~M. Czarnecki, V.~Zambaldi, M.~Jaderberg,
  M.~Lanctot, N.~Sonnerat, J.~Z. Leibo, K.~Tuyls, and T.~Graepel,
  ``Value-decomposition networks for cooperative multi-agent learning based on
  team reward,'' in {\em Proceedings of the 17th International Conference on
  Autonomous Agents and MultiAgent Systems}, AAMAS '18, (Richland, SC),
  p.~2085–2087, International Foundation for Autonomous Agents and Multiagent
  Systems, 2018.

\bibitem{qmix}
T.~Rashid, M.~Samvelyan, C.~S. De~Witt, G.~Farquhar, J.~Foerster, and
  S.~Whiteson, ``Monotonic value function factorisation for deep multi-agent
  reinforcement learning,'' {\em J. Mach. Learn. Res.}, vol.~21, jan 2020.

\bibitem{pmlr-v97-son19a}
K.~Son, D.~Kim, W.~J. Kang, D.~E. Hostallero, and Y.~Yi, ``{QTRAN}: Learning to
  factorize with transformation for cooperative multi-agent reinforcement
  learning,'' in {\em Proceedings of the 36th International Conference on
  Machine Learning} (K.~Chaudhuri and R.~Salakhutdinov, eds.), vol.~97 of {\em
  Proceedings of Machine Learning Research}, pp.~5887--5896, PMLR, 09--15 Jun
  2019.

\bibitem{Guo_2022_CVPR}
J.~Guo, Y.~Chen, Y.~Hao, Z.~Yin, Y.~Yu, and S.~Li, ``Towards comprehensive
  testing on the robustness of cooperative multi-agent reinforcement
  learning,'' in {\em Proceedings of the IEEE/CVF Conference on Computer Vision
  and Pattern Recognition (CVPR) Workshops}, pp.~115--122, June 2022.

\bibitem{he2023robust}
S.~He, S.~Han, S.~Su, S.~Han, S.~Zou, and F.~Miao, ``Robust multi-agent
  reinforcement learning with state uncertainties,'' 2023.

\bibitem{9283830}
J.~Lin, K.~Dzeparoska, S.~Q. Zhang, A.~Leon-Garcia, and N.~Papernot, ``On the
  robustness of cooperative multi-agent reinforcement learning,'' in {\em 2020
  IEEE Security and Privacy Workshops (SPW)}, pp.~62--68, 2020.

\bibitem{zhou2022romfac}
Z.~Zhou and G.~Liu, ``Romfac: A robust mean-field actor-critic reinforcement
  learning against adversarial perturbations on states,'' {\em arXiv preprint
  arXiv:2205.07229}, 2022.

\bibitem{huang2017adversarial}
S.~Huang, N.~Papernot, I.~Goodfellow, Y.~Duan, and P.~Abbeel, ``Adversarial
  attacks on neural network policies,'' {\em arXiv preprint arXiv:1702.02284},
  2017.

\bibitem{NEURIPS2020_f0eb6568}
H.~Zhang, H.~Chen, C.~Xiao, B.~Li, M.~Liu, D.~Boning, and C.-J. Hsieh, ``Robust
  deep reinforcement learning against adversarial perturbations on state
  observations,'' in {\em Advances in Neural Information Processing Systems}
  (H.~Larochelle, M.~Ranzato, R.~Hadsell, M.~Balcan, and H.~Lin, eds.),
  vol.~33, pp.~21024--21037, Curran Associates, Inc., 2020.

\bibitem{zhang2021robust}
H.~Zhang, H.~Chen, D.~S. Boning, and C.-J. Hsieh, ``Robust reinforcement
  learning on state observations with learned optimal adversary,'' in {\em
  International Conference on Learning Representations}, 2021.

\bibitem{han2022solution}
S.~Han, S.~Su, S.~He, S.~Han, H.~Yang, and F.~Miao, ``What is the solution for
  state adversarial multi-agent reinforcement learning?,'' {\em arXiv preprint
  arXiv:2212.02705}, 2022.

\bibitem{li2023attacking}
S.~Li, J.~Guo, J.~Xiu, P.~Feng, X.~Yu, J.~Wang, A.~Liu, W.~Wu, and X.~Liu,
  ``Attacking cooperative multi-agent reinforcement learning by adversarial
  minority influence,'' {\em arXiv preprint arXiv:2302.03322}, 2023.

\bibitem{oliehoek2016concise}
F.~A. Oliehoek and C.~Amato, {\em A concise introduction to decentralized
  POMDPs}.
\newblock Springer, 2016.

\bibitem{bernstein2002complexity}
D.~S. Bernstein, R.~Givan, N.~Immerman, and S.~Zilberstein, ``The complexity of
  decentralized control of markov decision processes,'' {\em Mathematics of
  operations research}, vol.~27, no.~4, pp.~819--840, 2002.

\bibitem{DE}
R.~Storn and K.~Price, ``Differential evolution – a simple and efficient
  heuristic for global optimization over continuous spaces,'' {\em J. of Global
  Optimization}, vol.~11, p.~341–359, dec 1997.

\bibitem{BILAL2020103479}
Bilal, M.~Pant, H.~Zaheer, L.~Garcia-Hernandez, and A.~Abraham, ``Differential
  evolution: A review of more than two decades of research,'' {\em Engineering
  Applications of Artificial Intelligence}, vol.~90, p.~103479, 2020.

\bibitem{OPARA2019546}
K.~R. Opara and J.~Arabas, ``Differential evolution: A survey of theoretical
  analyses,'' {\em Swarm and Evolutionary Computation}, vol.~44, pp.~546--558,
  2019.

\bibitem{8601309}
J.~Su, D.~V. Vargas, and K.~Sakurai, ``One pixel attack for fooling deep neural
  networks,'' {\em IEEE Transactions on Evolutionary Computation}, vol.~23,
  no.~5, pp.~828--841, 2019.

\bibitem{rummery1994line}
{\em On-line Q-learning using connectionist systems}, vol.~37.
\newblock 1994.

\bibitem{goodfellow2014explaining}
I.~J. Goodfellow, J.~Shlens, and C.~Szegedy, ``Explaining and harnessing
  adversarial examples,'' {\em arXiv preprint arXiv:1412.6572}, 2014.

\bibitem{Li_2020_CVPR}
M.~Li, C.~Deng, T.~Li, J.~Yan, X.~Gao, and H.~Huang, ``Towards transferable
  targeted attack,'' in {\em Proceedings of the IEEE/CVF Conference on Computer
  Vision and Pattern Recognition (CVPR)}, June 2020.

\bibitem{madry2018towards}
A.~Madry, A.~Makelov, L.~Schmidt, D.~Tsipras, and A.~Vladu, ``Towards deep
  learning models resistant to adversarial attacks,'' in {\em 6th International
  Conference on Learning Representations, {ICLR} 2018}, Conference Track
  Proceedings, (Vancouver, BC, Canada), OpenReview.net, 30 Apr -- 3 May 2018.

\bibitem{10.5555/3306127.3332052}
M.~Samvelyan, T.~Rashid, C.~Schroeder~de Witt, G.~Farquhar, N.~Nardelli,
  T.~G.~J. Rudner, C.-M. Hung, P.~H.~S. Torr, J.~Foerster, and S.~Whiteson,
  ``The starcraft multi-agent challenge,'' in {\em Proceedings of the 18th
  International Conference on Autonomous Agents and MultiAgent Systems}, AAMAS
  '19, (Richland, SC), p.~2186–2188, International Foundation for Autonomous
  Agents and Multiagent Systems, 2019.

\bibitem{schulman2017proximal}
J.~Schulman, F.~Wolski, P.~Dhariwal, A.~Radford, and O.~Klimov, ``Proximal
  policy optimization algorithms,'' {\em arXiv preprint arXiv:1707.06347},
  2017.

\bibitem{NEURIPS2022_9c1535a0}
C.~Yu, A.~Velu, E.~Vinitsky, J.~Gao, Y.~Wang, A.~Bayen, and Y.~WU, ``The
  surprising effectiveness of ppo in cooperative multi-agent games,'' in {\em
  Advances in Neural Information Processing Systems} (S.~Koyejo, S.~Mohamed,
  A.~Agarwal, D.~Belgrave, K.~Cho, and A.~Oh, eds.), vol.~35, pp.~24611--24624,
  Curran Associates, Inc., 2022.

\bibitem{sun2022who}
Y.~Sun, R.~Zheng, Y.~Liang, and F.~Huang, ``Who is the strongest enemy? towards
  optimal and efficient evasion attacks in deep {RL},'' in {\em International
  Conference on Learning Representations}, 2022.

\bibitem{9536399}
I.~Ilahi, M.~Usama, J.~Qadir, M.~U. Janjua, A.~Al-Fuqaha, D.~T. Hoang, and
  D.~Niyato, ``Challenges and countermeasures for adversarial attacks on deep
  reinforcement learning,'' {\em IEEE Transactions on Artificial Intelligence},
  vol.~3, no.~2, pp.~90--109, 2022.

\bibitem{10.5555/3237383.3238064}
A.~Pattanaik, Z.~Tang, S.~Liu, G.~Bommannan, and G.~Chowdhary, ``Robust deep
  reinforcement learning with adversarial attacks,'' in {\em Proceedings of the
  17th International Conference on Autonomous Agents and MultiAgent Systems},
  AAMAS '18, (Richland, SC), p.~2040–2042, International Foundation for
  Autonomous Agents and Multiagent Systems, 2018.

\bibitem{Weng2020Toward}
T.-W. Weng, K.~D. Dvijotham*, J.~Uesato*, K.~Xiao*, S.~Gowal*, R.~Stanforth*,
  and P.~Kohli, ``Toward evaluating robustness of deep reinforcement learning
  with continuous control,'' in {\em International Conference on Learning
  Representations}, 2020.

\bibitem{pham2022evaluating}
N.~H. Pham, L.~M. Nguyen, J.~Chen, H.~T. Lam, S.~Das, and T.-W. Weng,
  ``Evaluating robustness of cooperative marl: A model-based approach,'' {\em
  arXiv preprint arXiv:2202.03558}, 2022.

\bibitem{NIPS2017_68a97503}
R.~Lowe, Y.~WU, A.~Tamar, J.~Harb, O.~Pieter~Abbeel, and I.~Mordatch,
  ``Multi-agent actor-critic for mixed cooperative-competitive environments,''
  in {\em Advances in Neural Information Processing Systems} (I.~Guyon, U.~V.
  Luxburg, S.~Bengio, H.~Wallach, R.~Fergus, S.~Vishwanathan, and R.~Garnett,
  eds.), vol.~30, Curran Associates, Inc., 2017.

\bibitem{pmlr-v97-iqbal19a}
S.~Iqbal and F.~Sha, ``Actor-attention-critic for multi-agent reinforcement
  learning,'' in {\em Proceedings of the 36th International Conference on
  Machine Learning} (K.~Chaudhuri and R.~Salakhutdinov, eds.), vol.~97 of {\em
  Proceedings of Machine Learning Research}, pp.~2961--2970, PMLR, 09--15 Jun
  2019.

\end{thebibliography}

\end{document}